\documentclass[final,5p,times,twocolumn]{elsarticle}

\usepackage{amsmath}
\usepackage{mathtools}
\usepackage{xcolor}
\usepackage{enumitem}
\usepackage{comment}
\usepackage{graphicx}
\usepackage{multirow}
\usepackage{array}
\usepackage{caption}
\usepackage{fontawesome5}
\usepackage{hyperref}
\usepackage{wrapfig}
\usepackage{graphicx}
\usepackage{caption}
\usepackage{booktabs}
\usepackage{adjustbox}
\usepackage{footnote}
\makesavenoteenv{table}
\usepackage{caption}
\usepackage{amssymb}
\usepackage{booktabs}
\usepackage{titlesec}
\titlespacing\section{0pt}{10pt plus 4pt minus 2pt}{0pt plus 2pt minus 2pt}
\titlespacing\subsection{0pt}{8pt plus 4pt minus 2pt}{0pt plus 2pt minus 2pt}
\begin{document}

\begin{frontmatter}



\title{Evaluating the Effectiveness of XAI Techniques for Encoder-Based Language Models}

\author[1]{Melkamu Abay Mersha} 
\author[2]{Mesay Gemeda Yigezu}
% \author[1]{Joseph Wood}  
\author[1]{Jugal Kalita}
% \author[1,3]{Melkamu Mersha, Jugal Kalita } 

\affiliation[1]{organization={College of Engineering and Applied Science, University of Colorado Colorado Springs},
            addressline={},
            postcode={80918},
            state={CO},
            country={USA}}

\affiliation[2]{organization={Instituto Politécnico Nacional (IPN), Centro de Investigación en Computación (CIC)},
            addressline={},
            postcode={07738},
            state={Mexico city},
            country={Mexico}}

% % Authors from the first affiliation
% \author[1]{Melkamu Mersha}
% \author[1]{Joseph Wood}
% \author[1]{Ali Al Shami}
% \author[1]{Jugal Kalita}

% % Author from the second affiliation
% \author[2]{Khang Lam}

% % First affiliation
% \address[1]{College of Engineering and Applied Science, University of Colorado Colorado Springs, Colorado Springs, CO, USA}

% % Second affiliation
% \address[2]{College of Information and Communication Technology, Can Tho University, Can Tho, Vietnam}

\begin{abstract}
The black-box nature of large language models (LLMs) necessitates the development of eXplainable AI (XAI) techniques for transparency and trustworthiness. However, evaluating these techniques remains a challenge. This study presents a general evaluation framework using four key metrics: Human-reasoning Agreement (HA), Robustness, Consistency, and Contrastivity. We assess the effectiveness of six explainability techniques from five different XAI categories—model simplification (LIME), perturbation-based methods (SHAP), gradient-based approaches (InputXGradient, Grad-CAM), Layer-wise Relevance Propagation (LRP), and attention mechanisms-based explainability methods (Attention Mechanism Visualization, AMV)—across five encoder-based language models: TinyBERT, BERTbase, BERTlarge, XLM-R large, and DeBERTa-xlarge, using the IMDB Movie Reviews and Tweet Sentiment Extraction (TSE) datasets. Our findings show that the model simplification-based XAI method (LIME) consistently outperforms across multiple metrics and models, significantly excelling in HA with a score of 0.9685 on DeBERTa-xlarge, robustness, and consistency as the complexity of large language models increases. AMV demonstrates the best Robustness, with scores as low as 0.0020. It also excels in Consistency, achieving near-perfect scores of 0.9999 across all models. Regarding Contrastivity, LRP performs the best, particularly on more complex models, with scores up to 0.9371.
\end{abstract}

\begin{keyword}
XAI\sep explainable artificial intelligence\sep interpretable deep learning\sep explainable machine learning\sep evaluation framework\sep evaluation metrics\sep large language models\sep LLMs\sep interpretability\sep natural language processing\sep NLP\sep explainability techniques\sep black-box models.
\end{keyword}

\end{frontmatter}

\makeatletter
\def\ps@pprintTitle{%
    \let\@oddhead\@empty
    \let\@evenhead\@empty
    \def\@oddfoot{\hbox to \textwidth{\hfil\thepage\hfil}}%
    \let\@evenfoot\@oddfoot
}
\makeatother

\section{Introduction}
\label{section:introduction}
The exponential growth in the capabilities of powerful large language models (LLMs) such as GPT \cite{radford2018improving}, BERT \cite{devlin2018bert}, and their derivatives has revolutionized various domains, including safety-critical applications. However, these models are complex and opaque, posing significant challenges in understanding their decision-making processes and their internal workings. It is critically important to comprehend the underlying principles behind the decisions of these architecturally complex models. As these models become more sophisticated and are deployed across broader applications, the need for clear and interpretable explanations of their decision-making processes becomes increasingly essential for discovering potential flaws or biases \cite{das2020opportunities}, enhancing user trust \cite{arrieta2020explainable}, facilitating regulatory compliance \cite{regulation2016regulation}, and guiding the responsible integration of AI models into diverse sectors \cite{langer2021we}, \cite{arrieta2020explainable}.

Explainable Artificial Intelligence (XAI) techniques are the key to unlocking the reasons behind a model’s decision-making process. They provide crucial insights into how and why a model arrives at a particular decision, bridging the gap between complex AI models and human understanding \cite{shah2021neural}, \cite{dwivedi2023explainable}. Post-hoc XAI methods can be employed to analyze and interpret trained models, providing explanations for their decisions without modifying the underlying model structure. Based on their design and functionality, these post-hoc XAI techniques can be categorized into model simplification approaches, which create interpretable forms of complex models \cite{ribeiro2016should}; perturbation-based methods, which alter inputs to identify influential features \cite{zeiler2014visualizing}; gradient-based approaches, which use gradients to assess feature contributions \cite{sundararajan2017axiomatic}; Layer-wise Relevance Propagation (LRP) approaches, which trace decisions back through model layers to assign relevance scores \cite{bach2015pixel}; and attention mechanisms techniques, which highlight the key input features affecting model predictions based on the attention weights \cite{honnibal2017spacy}. In addition to these categories, several other post-hoc XAI methods exist, such as generating natural language explanations, which provide narrative interpretations of model decisions \cite{camburu2018snli}.

The growing number of explainability techniques in each XAI category often produce varying and sometimes contradictory explanations for the same input and model, complicating the determination of accuracy \cite{hassija2024interpreting}. Explanations may lack consistency across different situations due to changes in model behavior, making validation and verification challenging. Not all XAI methods suit every model architecture or complexity, highlighting the need for systematic assessment and automated selection of suitable techniques \cite{pawlicki2024evaluating}. Although previous studies have used various metrics to evaluate XAI effectiveness \cite{deyoung2019eraser}, inconsistencies among these studies result in a lack of standardized criteria for comparison \cite{pawlicki2024evaluating}. Additionally, these studies often overlook crucial factors like model complexity, language diversity, and application domain \cite{atanasova2024diagnostic, lukas2024bridging}, and most current metrics emphasize feature salience scores, potentially leading to misleading results when identical scores are assigned to different features.

In our study, we introduce new text ranking-based metrics alongside saliency scores and develop a robust evaluation framework that integrates the strengths of existing methods while addressing their limitations. This framework focuses on critical factors such as model complexity, a broad range of downstream tasks, diverse explainability categories, and the dynamic behavior of models over time. Unlike most existing studies, which primarily focus on a limited set of aspects, our approach comprehensively considers various explainability methods, downstream tasks, model architectures, language diversity, and domain-specific requirements to ensure a meaningful evaluation of XAI techniques.

We establish a general evaluation framework to assess XAI methods across different AI models, proposing metrics that account for the strengths and limitations of XAI methods in encoder-based language models. The framework includes widely applicable XAI methods from five distinct categories, evaluated on five models with varying complexities across two text classification tasks—short and long text inputs. Our evaluation, conducted on five encoder-only language models—TinyBERT, BERTbase, BERTlarge, XLM-R large, and DeBERTa-xlarge—provides a nuanced understanding of how different XAI techniques perform across various models and scenarios. The framework and metrics are designed to be applicable to a range of XAI techniques, encoder-based language models, and downstream tasks.

The contributions of this study are:

\begin{itemize} 
\setlength{\itemsep}{0pt}
  \setlength{\parskip}{0pt}
    \item We propose a comprehensive end-to-end evaluation framework to assess the effectiveness of XAI techniques across various machine learning, including transformer-based models. 
    \item We develop four evaluation metrics based on saliency token ranking and saliency score approaches for evaluating XAI techniques in encoder-based language models. 
    \item We analyze and compare five XAI categories across five encoder-based language models using these metrics on two downstream tasks. 
    \item We evaluate our XAI framework on two text classification tasks under five XAI categories and five encoder-based language models. 
    \item We compare and contrast explanations with human rationales to assess alignment with models' decision-making processes. 
    \item We provide comparative analyses to guide the selection of suitable XAI techniques for different encoder-based language models in downstream tasks. 
\end{itemize}

The remainder of this paper is organized as follows: Section 2 reviews the related work on XAI techniques and evaluation methods. In Section 3, we present the methodology, detailing the proposed evaluation framework and metrics. Section 4 describes the experimental setup, including datasets, models, and the selected XAI techniques. In Section 5, we discuss the results of the evaluation, providing insights into the effectiveness of different XAI methods across various encoder-based language models. Section 6 presents the limitations and future work. Finally, Section 7 concludes the paper, summarizing our findings and suggesting directions for future research.

\section{Background and Related Works}
\label{section:relatedwork}
\subsection{Background}
Explainable AI (XAI) has become essential to artificial intelligence and machine learning, especially with the rise of complex models like Large Language Models such as GPT \cite{radford2018improving}, BERT \cite{devlin2018bert}, and various transformer-based architectures. These models have performed exceptionally well in diverse natural language processing tasks. However, their complexity makes them operate largely as black boxes, presenting considerable challenges in understanding their internal decision-making processes. This opacity has critical implications for trustworthiness, fairness, and accountability, especially in high-stakes domains such as healthcare, finance, and law. To address these issues, XAI techniques have been developed to provide insights into AI models’ decision-making processes, offering explanations that enhance transparency and trust.

The urgency for evaluating XAI techniques on LLMs stems from the need to ensure that explanations are accurate and practically useful across different languages and NLP tasks. However, most existing XAI techniques were not developed with LLMs in mind. Consequently, they may struggle to provide coherent and comprehensive interpretations for these complex architectures, potentially requiring adaptations or entirely new methodologies that can handle the unique requirements of LLM interpretability. The need for new methodologies to handle the unique requirements of LLM interpretability is urgent and cannot be overstated.

\subsection{XAI Techniques}
XAI techniques are generally categorized into two categories: ante-hoc methods, which aim to be applied during model training, and post-hoc methods, which are applied after deployment to explain model predictions. Given LLMs' complexity, post-hoc techniques are often preferred as they offer flexibility for explaining pre-trained models.
In post-hoc explainability, methods are further divided into model-agnostic techniques, which can be applied to any black-box model, and model-specific techniques, designed for specific architectures.

Post-hoc explainability techniques are broadly categorized into five main groups based on their functionalities and design methodologies.
\textit{Model simplification techniques}, such as LIME \cite{ribeiro2016should}, are model-agnostic, and they simplify complex models into more interpretable ones. \textit{Perturbation-based techniques}, like SHAP \cite{shapley1953value}, are model-agnostic and modify feature values to measure their impact on predictions. 
\textit{Gradient-based techniques}, such as Integrated Gradients \cite{sundararajan2017axiomatic}, Grad-CAM \cite{selvaraju2017grad}, and Saliency Maps \cite{simonyan2013deep}, explain models by analyzing gradients. Gradient-based XAI methods for Transformer models in NLP tasks analyze the influence of input tokens on predictions by computing gradients of the model's output with respect to input tokens \cite{barkan2021grad}, \cite{fantozzi2024explainability}, \cite{wang2024gradient}. 
\textit{Layer-wise Relevance Propagation (LRP) techniques} assign predictions to input features by redistributing scores backward through the model’s layers \cite{montavon2017explaining}, \cite{achtibat2024attnlrp}. LRP, originally developed for image processing tasks, has been successfully adapted to NLP tasks to enhance interpretability in models like BERT. In the context of NLP, LRP identifies the contributions of individual input tokens to a model's final prediction by tracing their influence through the model's layers \cite{ali2022xai}, \cite{yang2018explaining}, \cite{fantozzi2024explainability}.
\textit{Attention mechanism techniques}, such as  Attention Rollout \cite{karras2020analyzing} and Attention Mechanism Visualization \cite{honnibal2017spacy}, visualize influential input features by analyzing attention weights.

\subsection{Related Works}
Various XAI methods have been proposed to improve interpretability in AI models, yet their effectiveness varies widely across models, requiring systematic assessment. Some approaches, like counterfactual explanations, aim to expose model reasoning by revealing instances that would lead to different predictions \cite{kindermans2019reliability, wachter2017counterfactual}. However, these methods often fall short in providing holistic model explanations \cite{jacovi2020towards}. Other studies have relied on human evaluation to assess the quality of XAI outputs, but this is often subjective and can vary significantly between users \cite{bansal2021does, lertvittayakumjorn2019human}.
Alternative evaluation approaches, such as Ground Truth Correlation, have emerged, comparing human-identified salient features with those generated by XAI methods \cite{arras2022clevr, deyoung2019eraser}. Another method evaluates the sufficiency of explanations by removing the most salient tokens identified by an XAI technique and observing the impact on model performance \cite{deyoung2019eraser}. While insightful, such metrics may not fully capture an XAI method’s effectiveness for highly parameterized LLMs due to their limited scalability and focus on simpler models like random forests and SVMs \cite{arreche2024xai}. Studies on CNN, LSTM, and BERT have further evaluated model simplification, perturbation, and gradient-based techniques, demonstrating their interpretative potential \cite{atanasova2024diagnostic}.

Most existing studies on evaluating explainability techniques vary widely in scope, covering different model architectures, methods, and tasks. Some focus on single architectures with multiple explainability methods \cite{deyoung2019eraser}, while others examine a range of models like CNN, LSTM, and BERT, using various techniques such as model simplification, perturbation, and gradient-based approaches \cite{atanasova2024diagnostic}, \cite{mersha2024explainability}. Many studies are also dataset-specific \cite{guan2019towards}, limiting the generalizability of their findings across domains and languages. This emphasis on high-resource languages highlights a gap in understanding XAI techniques for under-resourced languages, impacting the fairness and accessibility of AI globally. Additionally, most research overlooks the influence of model complexity on the effectiveness of explainability methods. Transformer models, for instance, vary significantly in complexity, from millions to billions of parameters \cite{ yigezu2024ethio}, \cite{tonja2023first}. Current research does not sufficiently explore how explainability methods perform across these complexities or which methods are best suited for different levels of model complexity.

These limitations highlight the necessity of our comprehensive study, which systematically evaluates the effectiveness of XAI methods across LLMs with varying complexities. Our research addresses gaps in previous studies by integrating four robust evaluation metrics to thoroughly assess how different levels of encoder-based language model complexity influence the performance of XAI techniques. By adopting this holistic approach, we aim to identify and recommend the most effective XAI categories and techniques tailored to specific encoder-based language model complexities, ensuring their optimal application across diverse models and tasks.

\section{Methodology}
\label{section:methodology}
\begin{figure}[ht]
\centering
\includegraphics[width=\linewidth]{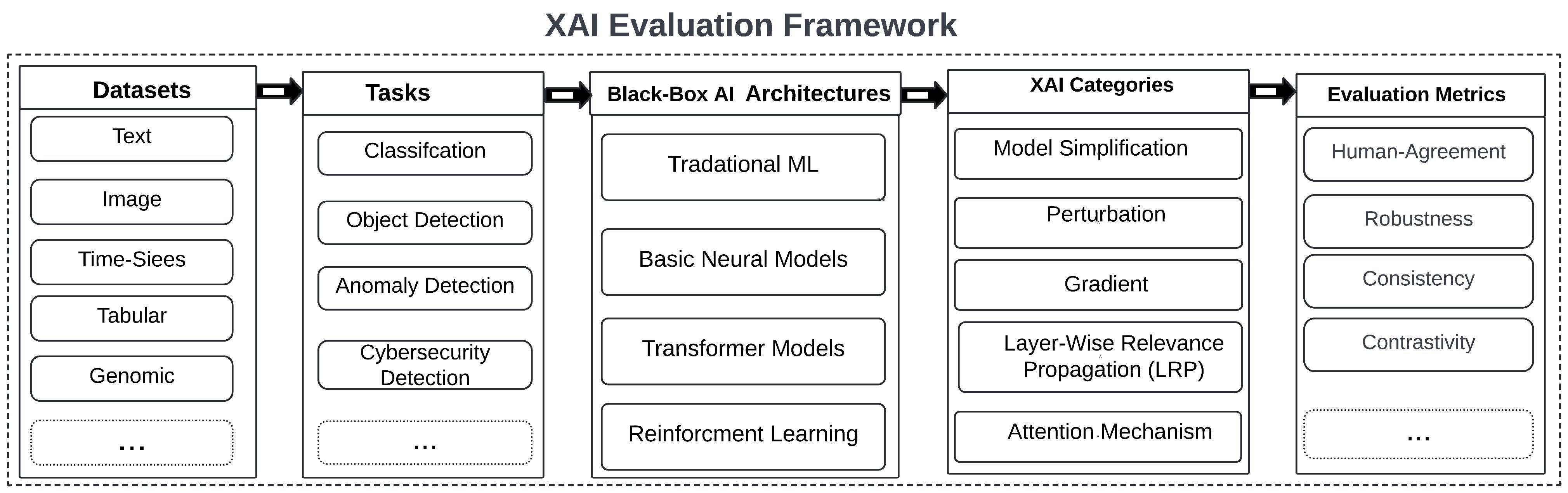}
\caption{An overview of our comprehensive XAI evaluation framework for assessing the effectiveness of explainability techniques across different scenarios.}
\label{fig:Evaluation_framework}
\end{figure}

An XAI evaluation framework has manifold applications, offering significant benefits across various AI models and XAI techniques development and deployment dimensions. Existing XAI evaluation frameworks are often simple and typically focus on limited tasks, a narrow range of AI models and explainability techniques, and evaluation metrics \cite{arreche2024xai}, \cite{atanasova2024diagnostic}, \cite{deyoung2019eraser}. These frameworks lack a comprehensive and standardized approach for evaluating the effectiveness of XAI methods across diverse contexts and applications. No general and standardized evaluation framework systematically assesses the performance of XAI techniques in a way that meets the needs of different stakeholders. Our new comprehensive XAI evaluation framework overcomes these limitations by incorporating a diverse array of datasets, covering a wide range of tasks, supporting various AI architectures, including neural networks and transformer models, and employing a variety of XAI methods and various evaluation metrics; see Figure ~\ref{fig:Evaluation_framework}. This comprehensive framework is used to rigorously evaluate and compare XAI methods across different scenarios.  The framework is designed to be easy to understand and extend, allowing for the incorporation of new datasets, tasks, AI models, XAI methods, and evaluation metrics as the field of XAI evolves. This holistic approach makes our framework exceptionally suited for in-depth evaluations of XAI techniques.

\subsection{Evaluation Metrics}
Evaluation metrics for XAI techniques are crucial as they provide quantitative measures to assess the quality and reliability of explanations \cite{nauta2023anecdotal}, \cite{mersha2024explainable}. These metrics ensure that explanations accurately reflect the model’s behavior. They also enable the systematic comparison and improvement of XAI methods, ensuring that AI models’ decisions are transparent and reliable across various applications. We build a comprehensive evaluation framework to evaluate various XAI categories by adopting and enhancing the existing metrics, and introducing new metrics.

\subsubsection{Human-reasoning Agreement (HA)}
The HA metric measures the degree of alignment between the explanations provided by an explainability technique and human intuition or reasoning \cite{deyoung2019eraser}, \cite{atanasova2024diagnostic}. It evaluates how closely a model's reasoning or explanation matches human judgment, with a human-annotated dataset serving as the baseline for this metric. However, it is important to note that the assumptions regarding the high degree of agreement between the feature importance scores (such as word saliency scores in this study) provided by explainability techniques and those from human-annotated datasets are not always valid. For instance, while the saliency scores for words may be similar, the specific words identified as salient may differ, which is a significant limitation of previous studies \cite{atanasova2024diagnostic}. Using measurements like cosine similarity, Pearson correlation, and intersection-over-union to compute the agreement between human and explainability word saliency scores with this limitation may not always be practical. This saliency approach does not adequately reflect the rank or relevance of words to the decision-making process. To address this limitation, we employed token/word ranking and Mean Average Precision (MAP) methods to assess the level of agreement between human-annotated and explainability-based explanations. Initially, we ranked the significant tokens/words for the model's decision-making process based on saliency scores from both perspectives to compute the Average Precision (AP) and MAP.

\textbf{ \textit{Mean Average Precision (MAP) }}
During our evaluation, we utilize the AP and MAP of the ranked words/tokens to precisely measure the agreement between human rationales and the explanations generated by the explainability technique and mathematically represented by Equations \ref{equ.AP1} and \ref{equ.MAP}, respectively, this metric provides a clear understanding of the alignment between human-annotated and explainability explanations. Average precision for a single instance evaluates the importance of salient words identified in the explanation compared to those identified by human annotators, aiding decision-making processes. It is computed as:
\begin{equation} \label{equ.AP1}
AP = \frac{\sum_{k=1}^{n} (P(k) \times rel(k))}{\text{Number of relevant tokens (n)}} 
\end{equation}
Where \( k \) represents the rank or position of a word in the sequence of retrieved words, \( n \) is the total number of retrieved words, \( P(k) \) is the precision at rank \( k \) in the ranked word list, and \( \text{rel}(k) \) is a binary indicator function. \( \text{rel}(k) = 1 \) if the word at rank or position \( k \) from the explainability explanation matches the corresponding word in the human annotation, otherwise \( \text{rel}(k) = 0 \), Equation \ref{equ.relevance}. The ranking of words from the explainability method is automatically determined based on relevance scores computed by the explainability technique. In contrast, human annotators provide a ranked order of words directly in their explanation rather than assigning explicit relevance scores only.  An example is available in \ref{Example}.\\
AP measures the precision of the explanation provided by the explainability technique for a single instance. If the AP score is high (closer to 1), it indicates that the explainability of that instance strongly aligns with the human rationale. A lower AP score (closer to 0) suggests that the agreement is poor in that particular instance.

MAP is the mean of the AP scores for all instances, as calculated by Equation (\ref{equ.MAP}).
\begin{equation}
\label{equ.MAP}
MAP = \frac{\sum_{n=1}^{N} AP_n}{N} 
\end{equation}
where $N$ is the total number of instances evaluated, $AP_{n}$ is the Average Precision calculated for the $n^{th}$ instance.

MAP evaluates the alignment between explanations provided by an explainability technique and human rationale across a set of instances or documents rather than measuring the precision of the explanation for each individual instance. A high MAP score (closer to 1) indicates that the explainability technique consistently performs well across diverse instances and strongly agrees with human rationales. Conversely, a low MAP score (closer to 0) suggests that the explainability technique's performance varies across instances, indicating poor agreement with human rationales.

\subsubsection{Robustness}
This metric measures the robustness of explanations provided by explainability techniques in response to changes in the input data and the model \cite{rosenfeld2021better}. It evaluates how explanations vary under various conditions, such as real-time applications and adversarial perturbations. It also assesses the model's consistency when its parameters are altered or when it is retrained with modified or augmented datasets. This helps understand the robustness and reliability of the explanations across different scenarios, thereby enhancing comprehension of the model’s behavior under diverse conditions \cite{nogueira2018stability}. 
We assess how stable the explanation provided for an original input instance $X$ remains when the input is slightly modified to \( X' \). In previous studies, the robustness of explanations is measured directly by evaluating the differences in top-k saliency token/word scores between the original input instance and the perturbed or modified input instance \cite{arreche2024xai}. However, this approach does not fully capture the robustness of explanations or the model’s behavior, as different tokens/words at the same rank may have similar scores. To address this, we employ element-wise differences and averaging techniques to quantify robustness metrics at both the token/word and instance levels based on relevance computation.

\textbf{Relevance Function:} The relevance function, $rel(k)$, serves as a binary indicator to determine the relevance of each word $k$ in a model's decision-making process. If the word $k$ is included in both $X $ and $X'$, it returns 1; otherwise, it returns 0.

\begin{equation} 
\label{equ.relevance}
\text{rel}(k) = 
\begin{cases} 
1 & \text{if } k \in X \\
0 & \text{otherwise}
\end{cases}
\end{equation}
where $k$  is a relevant word to the model's decision-making process.\\
To create a modified instance \( X' \), a perturbation $\delta_{i}$ is applied such that \(X' \)=$X$ +$\delta_{i}$.  This perturbation $\delta_{i}$ typically involves various techniques such as masking, replacing words with synonyms, removing words, or applying other modifications to words with high or low salience scores.

\textbf{Element-wise Difference $d(k)$:} For each word $k$, the function $d(k)$ precisely quantifies the change in the saliency scores of a word k between $X$ and \( X' \), as identified by $rel(k)$. $ d (k) $ is computed as: 

\begin{equation}
\label{equ.elementWise}
% d(k) = X[k] - X'[k] \cdot \text{rel}(k)
d(k) = \|X[k] - \left ( X'[k] \times {rel}(k) \right) \| 
\end{equation}
 where $X[k]$ represents the saliency score of word $k$ in the explanation derived from the original input, while $X'[k]$ represents the saliency score of word $k$ in the explanation derived from the modified input. The product with $rel(k)$ ensures that differences are calculated only for words that appear in both sets ($X$ and \( X' \)), avoiding distortion from irrelevant words.
 
\textbf{Average Difference (AD):} The AD aggregates the individual differences $d(k)$ for all relevant words and provides a single metric for each instance. AD reflects the average magnitude of change in an individual explanation due to an input modification. AD is essential for understanding the robustness of explanations at an instance level. Mathematically, it is described by Equation \ref{equ.AD}.

\begin{equation}
\label{equ.AD}
\text{AD} = \frac{1}{K} \sum_{k =1} ^{K} d(k)
\end{equation}
where $K$ represents the total number of words in an explanation for a given instance.

\textbf{Mean Average Difference (MAD):} MAD is a dataset-wide metric that averages the AD values across all instances, providing a global measure of the robustness of the explanations in response to changes in input data throughout the dataset. Mathematically, it is represented by Equation \ref{equ.MAD}
\begin{equation}
% \label{equ.MAD}
\text{MAD} = \frac{1}{N} \sum_{n =1} ^{N} AD
\end{equation}

\begin{equation}
\label{equ.MAD}
MAD = \frac{ \sum_{n=1}^N \left( \frac{1}{K} \sum_{k=1}^K d(k) \right) }{N}
\end{equation}
where $N$ is the total number of instances.\\
Lower AD and MAD scores indicate that the explainability technique performs robustly well both at the instance level and across diverse instances, respectively.

\subsubsection{Consistency}
Models with diverse architectures tend to have low explanation consistency and vice versa \cite{huang2022conceptexplainer}. We are interested in proving the similarity of attention reasoning mechanisms rather than similar predictions for similar model architecture. We focus on a set of models with the same architecture, trained with different random seeds and randomly initialized weights. Our interest is to discover the attention-based reasoning mechanisms instead of model prediction outputs since similarities of prediction outputs are not always guaranteed in models with similar reasoning. Different models can arrive at the same prediction through different reasoning processes. \\
Let $M_{a }$ and $M_{b}$ be two distinct models with similar architectures and trained with different seeds, and $x_{i}$ be an input instance. $D_A(M_a, M_b, x_i)$ and $ D_E (M_a, M_b, x_i) $ are the distances between attention weights and explanation scores, respectively, and they can be computed using some distance measurements such as Cosine similarity or Euclidean distance.

\textbf{Multi-Layer Attention Weights Distance for input $x_i$}:\\
% Assume we can have two models ($M_a$ and $M_b$) with similar architectures and input $x_i$. 
The similarity of their reasoning mechanisms can be effectively measured by computing the distance between the attention weights generated by the two models (Equation \ref{equ.MeanAverageAttention}). First, we extracted and averaged attention weights. For a model $M$ with $L$ attention layers, attention weights  at $l^{th}$ layer can be represented by $A_l$($M$, $x_i$) for the input $x_i$, ($M $ can be  model $M_a$ or $M_b$). Then, we compute the weighted average of attention weights by Equation \ref{equ.AverageAttention}.

% \textbf{Attention Distance for Input $x_{i}$:}

\begin{equation}
\label{equ.AverageAttention}
\overline{A}(M, x_i) = \frac{1}{L} \sum_{l=1}^L A_l(M, x_i) 
\end{equation}
Here, $\overline{A}(M, x_i)$ represents the averaged attention weights over all layers for a given instance $x_i$. We obtain the average attention weight vector $\overline{A}(M, x_i)$. This average provides a consolidated view of how the model attends to different parts of the input across all layers, making it easier to compare attention mechanisms between models.

The distance between model $M_a$ and $M_b$, denoted as ($D_{A}(M_a, M_b, x_i)$) is computed using the equation below.

\begin{equation}
\label{equ.distanMaMb}
D_{A}(M_a, M_b, x_i) = D_{A}(M_a(x_i), M_b(x_i))
\end{equation}
% where $D_{A}(M_a, M_b, x_i)$ is the distance between model $M_a$ and $M_b$.\\
$D_{A}(M_a, M_b, x_i)$ establishes a way to quantify the differences or similarities in how two models process the same input $x_i$. This sets the basis for comparing models based on their responses to the same data point rather than only on their output predictions. $\overline{A}(M_a, x_i)$  and $\overline{A}(M_b, x_i)$ are the averaged attention weights of models $M_a$ and $M_b$ for the input $x_i$, as shown by Equation 
\ref{equ.AverageAttention}.Then $D_{A}(M_a, M_b, x_i)$ computed as:

\begin{equation}
\label{equ.MeanAverageAttention}
D_{A}(M_a, M_b, x_i) = D_{A}(\overline{A}(M_a, x_i), \overline{A}(M_b, x_i)).
\end{equation}

Considerable attention weight similarities or differences suggest that the models attend to different aspects of the input being trained with different seeds. This is not just a technical detail but a crucial aspect in assessing whether the models maintain consistent focus and importance on the same input features (words), which is our main interest in evaluating the consistency and effectiveness of different explainability techniques.

% Explanation scores
\textbf{Explanation Score Distance for Input $x_i$:}
The explanation scores are derived from an explainability technique that we are interested in to evaluate its effectiveness and consistency. By measuring the distance between the explanation scores, we can evaluate how similarly or differently the explainability techniques explain the two models' predictions for the same input.

\begin{equation}
\label{equ.explantion}
% D_E(M_a, M_b, x_i) = \| E(M_a, x_i) - E(M_b, x_i) \|
D_{E}(M_a, M_b, x_i) = D_{E}(M_a(x_i), M_b(x_i))
\end{equation}
This equation computes the difference in explanation scores between the two models for the same input, emphasizing the magnitude of differences.

\textbf{Consistency in Instance Level} $x_i$:
We can assess effectiveness and consistency at the instance level by comparing the distances between attention weights ($D_A(M_a, M_b, x_i)$) and explanation scores ($D_E(M_a, M_b, x_i)$) for individual inputs. If these distances are similar or close to similar, the explainability technique is considered effective, indicating that models with similar attention weights provide consistent explanations.

\textbf{Consistency across the Dataset:}
For multiple instances \( X = \{ x_1, x_2, \ldots, x_N \} \), we computed the correlation of the distances of attention weights and explanation scores using Spearman's rank correlation coefficient $\rho$ \cite{spearman1961proof}, (Equation \ref{equ.spearmancorreliation1}).

\begin{equation}
\label{equ.spearmancorreliation1}
\rho = \text{Spearman's} \, corr \left( \{ D_A(M_a, M_b, x_i) \}_{i=1}^N, \{ D_E(M_a, M_b, x_i) \}_{i=1}^N \right)
\end{equation}
where $\rho$ is the overall Spearman’s correlation of the attention weights and explanations of the entire input instances, and $N$ is the total number of input instances.\\
 $\rho $ provides the overall correlation between the attention weights and explainability explanations of the global trend across all data points. The higher the positive correlation, the more consistent the explainable technique is.
$\rho$  measures the strength and direction of the monotonic relationship between the distances in attention weights and the distances in explanation scores across multiple inputs. A high correlation indicates that models with similar attention weights also tend to have similar explanation scores, suggesting consistency in their reasoning mechanisms. 

By using this attention-weight approach, we can determine whether models trained with different random seeds exhibit consistent reasoning mechanisms and focus on similar input words, thereby evaluating the effectiveness and robustness of explainability techniques.

\subsubsection{Contrastivity}
Contrastivity is a critical evaluation metric for assessing the effectiveness of XAI methods, particularly in classification tasks \cite{stepin2021survey}. It focuses on how well an XAI method can differentiate between different classes through its explanations, providing insight into why a model chooses one class over another. For example, to assess the difference between the two classes (positive or negative), we can compare the explanations for different class predictions and see if the explanations for one class are distinct from those for another. This practical use of contrastivity helps us to understand the effectiveness of XAI methods. We used Kullback-Leibler Divergence (KL Divergence) to quantify the contrastivity metric in feature importance distributions. KL Divergence is ideal for its sensitivity to differences in feature importance distributions and its focus on the direction of divergence, making it perfect for analyzing and comparing tokens/words (feature) importance across different classes \cite{kullback1951information}.

\begin{equation}
    \text{KL}(P \parallel Q) = \sum_{i=1}^n P(i) \log \left( \frac{P(i)}{Q(i)} \right)
\end{equation}
where $P(i)$ and $Q(i)$ represent the importance of feature $i$ in one class and in a different class, respectively, and $n$ is the total number of tokens/words in the given instance. $P$ and $Q$ are the distribution of feature importance for the two different classes.

High contrastivity means that the XAI method effectively highlights different features for different classes.
The positive attributions should be associated with the target label, and the negative attributions should be associated with the opposite class.

\section{Experiments and Setups}
\label{section:experiments}
We have proposed a comprehensive XAI evaluation framework as a benchmark for assessing the effectiveness of explainability techniques across different scenarios (Fig.~\ref{fig:Evaluation_framework}). Our study used five different encoder-only language models with varied levels of complexity to focus on text data and a classification task. To provide clear insights into our experiment, we included six distinct XAI methods, each representing five different categories of XAI techniques (two methods for gradient-based categories). We then evaluated these methods using four specific metrics.

We conducted our experiments on Google Colab, utilizing an NVIDIA A100-SXM4-40GB GPU with 40 GB of VRAM powered by CUDA Version 12.2. This setup provided robust computational resources, enabling efficient handling of high-demand tasks such as deep learning model training and large-scale data processing.

\subsection{Datasets}
We utilized two distinct datasets for our study: IMDB \footnote{https://www.kaggle.com/datasets/columbine/imdb-dataset-sentiment-analysis-in-csv-format} Movie Reviews and Tweet Sentiment Extraction. The IMDB Movie Reviews dataset consists of 50,000 movie reviews, each labeled as either positive or negative. The Tweet Sentiment Extraction (TSE) \footnote{https://www.kaggle.com/c/tweet-sentiment-extraction} dataset consists of 31,016 tweets labeled with sentiments such as positive, negative, or neutral. Tweets are typically short texts. We randomly split each dataset into 80\% for training and 20\% for testing.

\subsection{Models}
\begin{figure}[ht]
\centering
\includegraphics[width=\linewidth]{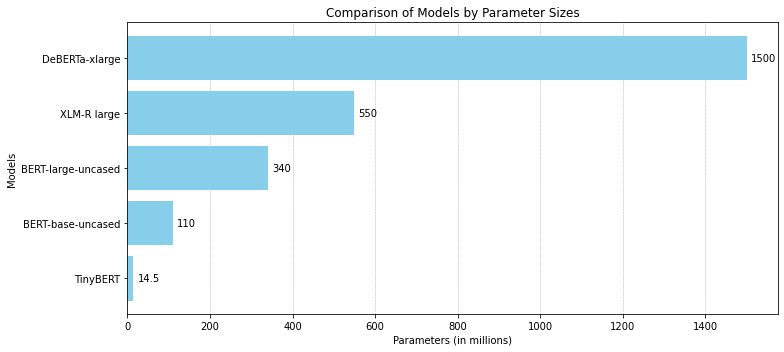}
\caption{The selected transformer-based models by complexity band parameter size.}
\label{fig:models}
\end{figure}

We conducted experiments using commonly used transformer-based models, including TinyBERT \cite{jiao2019tinybert}, BERT-base-uncased \cite{devlin2018bert}, BERT-large-uncased \cite{devlin2018bert}, XLM-R large \cite{conneau2019unsupervised}, and DeBERTa-xlarge \cite{he2020deberta}. These models were chosen primarily for their varying levels of complexity and parameter sizes. This baseline model selection enables a comprehensive comparison and analysis of explainability techniques across models with diverse complexities. Fig ~\ref{fig:models}  illustrates the selected transformer-based models and their respective parameter sizes, highlighting the differences that impact their performance and suitability for various tasks. By evaluating the effectiveness of explainability techniques, we aim to understand how model complexity and size influence the interpretability and transparency of these models in practical applications. We fine-tune all the selected pre-trained models by adding a linear layer on top of them. The size of this linear layer corresponds to the number of classes in the given classification task.

\subsection{Explainability Techniques}
We grouped explainability techniques into five categories based on their design principles and functionality to provide a comprehensive evaluation by category: model simplification \cite{ribeiro2016should}, perturbation \cite{zeiler2014visualizing}, gradient \cite{sundararajan2017axiomatic}, layer-wise relevance propagation \cite{bach2015pixel}, and attention mechanism \cite{honnibal2017spacy}.  We chose the most representative and commonly used explainability techniques from each category.

From the model simplification category, we selected LIME, a model-agnostic explainer. LIME trains a linear model to approximate the local decision boundary for each instance and provides explanations for individual predictions of a complex model   \cite{ribeiro2016should}.

We selected SHAP, another model-agnostic explainer, from the perturbation category. SHAP uses the perturbation technique to determine the importance of each feature and provides interpretable explanations for the model’s predictions 
\cite{sundararajan2017axiomatic}, \cite{shapley1953value}. 

For the Gradient category, we employed InputXGradient and Gradient-weighted Class Activation Mapping (Grad-CAM), model-specific techniques. These methods leverage the gradients of the model’s output with respect to its input features to interpret the predictions of deep learning models \cite{kindermans2017learning}, \cite{selvaraju2017grad}.

We used LRP-$\epsilon$ (LRP for simplicity for this study) methods from the Layer-wise Relevance Propagation category due to its balanced approach between simplicity and stability.  LRP-$\epsilon$ is a model-specific technique that assigns a model’s prediction to its input features by systematically redistributing the prediction score of the model backward through each neuron to the previous layer based on the contribution of each neuron to the output  \cite{bach2015pixel}, \cite{yang2018explaining}.

% \textit{Attention mechanism explainability techniques} are used to visualize and interpret the most influential input features for a model's prediction based on the attention weights assigned by the model's attention mechanisms
Lastly, we selected an Attention Mechanism Visualization (AMV) explainability technique that is also model-specific. This technique visualizes and interprets the most influential input features for a model’s prediction based on the attention weights assigned by the model’s attention mechanism 
\cite{honnibal2017spacy}, \cite{chefer2021transformer}, \cite{ali2022xai}.

\begin{table*}[ht]
\caption{The quantitative results of the Human-reasoning Agreement metric on various XAI methods and encoder-based language models for IMDB and TSE datasets. Higher scores indicate better agreement. }
\centering
\scriptsize
\begin{adjustbox}{max width=\textwidth}
\begin{tabular}{ccccccccccccccccccccccc}
\toprule
& \multicolumn{5}{c}{IMDB} & \multicolumn{5}{c}{TSE} \\
\cmidrule(lr){2-6} \cmidrule(lr){7-11}
 &TinyBERT  & BERTbase &  BERTlarge & XLM-R &  DeBERTa\_xlarge &TinyBERT & BERTbase & BERTlarge & XLM-R &  DeBERTa\_xlarge  \\
\midrule
LIME & \textbf{ 0.8774} & \textbf{0.6981 } &\textbf{0.8903}& \textbf{0.9445} & \textbf{0.9685} &\textbf{ 0.7566} & 0.4689 & \textbf{0.8023} & \textbf{0.8869} & \textbf{0.9118}  \\
SHAP & 0.4135 & 0.4354 &0.5012& 0.5634 & 0.6625 & 0.5231 & 0.5728 & 0.6002 & 0.6894 & 0.7254  \\
LRP & 0.6427 & 0.2736 & 0.2078 & 0.2011 & 0.1984 & 0.7223 & \textbf{0.5986 }& 0.5023 & 0.4533 & 0.3689 \\
 InputXGradient & 0.0782 & 0.0659 & 0.1775 & 0.2356 & 0.3133 & 0.1691 & 0.0965 & 0.2647 & 0.3562 & 0.3959 \\
Grad-CAM & 0.1437 & 0.1936 & 0.5229 & 0.6118 & 0.6497 & 0.2563 & 0.4125 & 0.4565 & 0.5001 & 0.5926 \\
AMV & 0.1658 & 0.1459 & 0.1001 & 0.0859 & 0.0653 & 0.2136 & 0.1759 & 0.1325 & 0.0962& 0.0593 \\
\bottomrule
\end{tabular}
\end{adjustbox}
\label{HA}
\end{table*}

\begin{table*}[htbp!]
\caption{The quantitative results of the Robustness metric on various XAI methods and encoder-based language models for IMDB and TSE datasets. Lower scores indicate better robustness.}
\centering
\scriptsize
\begin{adjustbox}{max width=\textwidth}
\begin{tabular}{ccccccccccccccccccccccc}
\toprule
& \multicolumn{5}{c}{IMDB} & \multicolumn{5}{c}{TSE} \\
\cmidrule(lr){2-6} \cmidrule(lr){7-11}
 &TinyBERT  & BERTbase &  BERTlarge & XLM-R &  DeBERTa\_xlarge &TinyBERT & BERTbase & BERTlarge & XLM-R &  DeBERTa\_xlarge  \\
\midrule
LIME & 0.0056 & 0.0058 & 0.0061 & 0.0078 & 0.0092 & 0.0043 & 0.0049& 0.0053 & 0.0060 & 0.0068 \\
SHAP & 0.0356 & 0.0387 &0.1258 & 0.1547 & 0.2139 & 0.0258 & 0.0301& 0.03662& 0.1321 & 0.1965  \\
LRP & 0.3214 & 0.5431 & 0.7621 & 0.8549 & 0.9124 & 0.3392 & 0.4895 & 0.5684 & 0.6855 & 0.8215 \\
InputXGradient & 0.1108 & 0.1546 & 0.2391 & 0.2769 & 0.3012 & 0.0953 & 0.1258 & 0.2011 & 0.2547 & 0.2958 \\
Grad-CAM & 0.0237 & 0.0161 & 0.0273 & 0.0312 & 0.0367 & 0.0189 & 0.0123 & 0.0232 & 0.0259 & 0.0291 \\
AMV & \textbf{0.0020} & \textbf{0.0023} & \textbf{0.0056} & \textbf{0.0058} & \textbf{0.0073} & \textbf{0.0014} & \textbf{0.0019} & \textbf{0.0032} & \textbf{0.0041} & \textbf{0.0051} \\
\bottomrule
\end{tabular}
\end{adjustbox}
\label{robustness}
\end{table*}

\section{Results and Discussion}
\label{section:results}
We present the results of each evaluation metric on various XAI methods and encoder-based language models across the IMDB (long texts) and TSE (short texts) datasets, considering model complexity from TinyBERT (14.5 million parameters) to DeBERTa-large (1.5 billion parameters). The primary focus of this study is not on the performance of the models themselves but on the effectiveness of the XAI methods on these encoder-based language models. The quantitative results for each evaluation metric are presented below. Additionally, \ref{Visualization} includes sample visualizations of the outputs from each XAI technique, offering deeper insights and enabling thorough comparisons.

\textbf{Human-reasoning Agreement (HA):}
Table ~\ref{HA} presents the HA metric results. We randomly selected 100 instances from each dataset. We used three machine learning experts to annotate the datasets as the baseline to evaluate the alignment between explanations provided by XAI methods and human judgment.

On the IMDB dataset, which contains longer texts, LIME consistently performs exceptionally well across all model sizes, with HA scores improving as model complexity increases. This suggests that LIME effectively captures human-like explanations irrespective of model size, achieving its highest score of 0.9685 with DeBERTa-xlarge. SHAP shows moderate performance, with its effectiveness significantly improving for larger models, indicating that it benefits from increased model complexity or larger parameter sizes. LRP struggles with larger models, displaying a decline in performance as model complexity increases. It performs better with simpler models like TinyBERT, where it achieved its highest score of 0.6427. InputXGradient, despite showing low agreement with human rationales, holds promise as it improves with larger models, suggesting that it benefits from increased model complexity. Grad-CAM demonstrates moderate performance, with improvements seen as model size increases, while AMV shows consistently poor performance across all models, with a slight decrease as model complexity increases.

For the TSE dataset, which contains shorter texts, LIME again shows high agreement with human rationales, with performance improving as model complexity increases, achieving a top score of 0.9118 with DeBERTa-xlarge. SHAP, demonstrating its adaptability, performs better on TSE compared to IMDB, particularly for larger models, indicating its effectiveness with increased model complexity. LRP performs better on TSE (short text) than IMDB but still shows a decline in performance with larger models. InputXGradient improves on TSE compared to IMDB, with better performance for larger models, while Grad-CAM maintains moderate performance, improving with larger models. AMV continues to show low performance across all models, with a slight decrease as model complexity increases.

% \textbf{Overall,} 
LIME stands out as the best-performing explainability technique, providing strong and reliable alignment with human rationales across both datasets and various models. Its performance has improved with increased model complexity. SHAP and Grad-CAM, on the other hand, provide a balance of performance, benefiting significantly from larger models. In contrast, AMV and LRP are the least effective in aligning with human rationales and become significantly less effective as model complexity increases. InputXGradient is also the least effective but benefits from increasing model complexity.

 \begin{table*}[htbp!]
\caption{The quantitative results of Consistency metric on various XAI methods and encoder-based language models for IMDB and TSE datasets. Higher scores indicate better consistency.}
\centering
\scriptsize
\begin{adjustbox}{max width=\textwidth}
\begin{tabular}{ccccccccccccccccccccccc}
\toprule
& \multicolumn{5}{c}{IMDB} & \multicolumn{5}{c}{TSE} \\
\cmidrule(lr){2-6} \cmidrule(lr){7-11}
 &TinyBERT  & BERTbase &  BERTlarge & XLM-R &  DeBERTa\_xlarge &TinyBERT & BERTbase & BERTlarge & XLM-R &  DeBERTa\_xlarge  \\
\midrule
LIME & 0.9665 & 0.9741 & 0.9800 & 0.9837 & 0.9895 & 0.7568& 0.8425 & 0.8962 & 0.9228& 0.9556 \\
SHAP & 0.9002 & 0.9368 &0.9487 & 0.9569 & 0.9775 & 0.8322 & 0.8598 & 0.8896& 0.9002& 0.09324  \\
LRP & 0.9417 & 0.9642 & 0.9754 & 0.9801& 0.9887 & 0.8556 & 0.8901 & 0.9223 & 0.9596 & 0.9713 \\
InputXGradient & 0.9341 & 0.9447 & 0.9555 & 0.9599 & 0.9799 & 0.7583 & 0.7759& 0.7952 & 0.8235& 0.8596 \\
Grad-CAM & -0.9593 & -0.9677 & -0.9698 & -0.9700 & -0.9749 & -0.8259 & -0.8556 & -0.8718 & -0.8987 & -0.9325 \\
AMV & \textbf{0.9999} & \textbf{0.9999} & \textbf{0.9999} & \textbf{0.9999} & \textbf{0.9999} & \textbf{0.9999} & \textbf{0.9999} & \textbf{0.9999} & \textbf{0.9999} & \textbf{0.9999} \\
\bottomrule
\end{tabular}
\end{adjustbox}
\label{consitency}
\end{table*}

\begin{table*}[htbp!]
\centering
\caption{The quantitative results of Contrastivity metric on various XAI methods and encoder-based language models for IMDB and TSE datasets. Higher scores indicate better.}
\scriptsize
\begin{adjustbox}{max width=\textwidth}
\begin{tabular}{ccccccccccccccccccccccc}
\toprule
& \multicolumn{5}{c}{IMDB} & \multicolumn{5}{c}{TSE} \\
\cmidrule(lr){2-6} \cmidrule(lr){7-11}
 & TinyBERT & BERTbase & BERTlarge & XLM-R & DeBERTa\_xlarge & TinyBERT & BERTbase & BERTlarge & XLM-R & DeBERTa\_xlarge \\
\midrule
LIME & 0.7065 & \textbf{0.7226} & 0.6545 & 0.6288 & 0.5766 & 0.6863 & \textbf{0.7145} & 0.6631 & 0.6251 & 0.5838 \\
SHAP & 0.6449 & 0.6694 & 0.6208 & 0.5565 & 0.5448 & 0.6552 & 0.6727 & 0.6075 & 0.5709 & 0.5356 \\
LRP & \textbf{0.7723} & 0.5797 & \textbf{0.7958} & \textbf{0.8456} & \textbf{0.9371} & \textbf{0.8598} & 0.7126 & \textbf{0.8540} & \textbf{0.8797} & \textbf{0.9367} \\
InputXGradient & 0.7488 & 0.5921 & 0.4493 & 0.4289 & 0.3942 & 0.7968 & 0.6833 & 0.5288 & 0.4516 & 0.3661 \\
Grad-CAM & 0.5689 & 0.4695 & 0.4163 & 0.3915 & 0.3641 & 0.6623 & 0.5965 & 0.4471 & 0.4034 & 0.3371 \\
AMV & 0.1546 & 0.1168 & 0.1012 & 0.0761 & 0.0384 & 0.1841 & 0.1616 & 0.1243 & 0.1050 & 0.0680 \\
\bottomrule
\end{tabular}
\end{adjustbox}
\label{contrastivity}
\end{table*}

\textbf{Robustness:}
Table ~\ref{robustness} presents the quantitative results of the robustness metric. 
LIME demonstrates very good robustness across all models for the IMDB dataset, with scores ranging from 0.0056 for TinyBERT to 0.0092 for DeBERTa-xlarge. This suggests that LIME's explanations are stable and consistent even for complex models handling long texts. SHAP, on the other hand, shows moderate robustness, with scores ranging from 0.0356 for TinyBERT to 0.2139 for DeBERTa-xlarge. This indicates that SHAP's explanations are less robust for larger models. LRP exhibits poor robustness, especially for larger models, with scores ranging from 0.3214 for TinyBERT to 0.9124 for DeBERTa-xlarge, suggesting highly variable and unstable explanations. InputXGradient shows moderate robustness, with scores ranging from 0.1108 for TinyBERT to 0.3012 for DeBERTa-xlarge, indicating that while it is more robust than LRP, it is less effective than LIME and Grad-CAM. Grad-CAM demonstrates good robustness, with scores ranging from 0.0161 for BERTbase to 0.0367 for DeBERTa-xlarge, providing relatively stable explanations even for larger models handling long texts. AMV shows the best robustness across all models, with scores ranging from 0.0020 for TinyBERT to 0.0073 for DeBERTa-xlarge, indicating highly stable and consistent explanations regardless of model complexity.\\
LIME maintains high robustness across different models for the TSE dataset, with scores ranging from 0.0043 for TinyBERT to 0.0068 for DeBERTa-xlarge, suggesting stable and consistent explanations for short texts. SHAP again demonstrates moderate robustness, with scores ranging from 0.0258 for TinyBERT to 0.1965 for DeBERTa-xlarge, indicating less stable explanations for larger models. LRP continues to show poor robustness, especially for larger models, with scores ranging from 0.3392 for TinyBERT to 0.8215 for DeBERTa-xlarge. InputXGradient shows moderate robustness, with scores ranging from 0.0953 for TinyBERT to 0.2958 for DeBERTa-xlarge, providing more stable explanations than LRP but less effective than LIME and Grad-CAM. Grad-CAM demonstrates good robustness, with scores ranging from 0.0123 for BERTbase to 0.0291 for DeBERTa-xlarge, providing relatively stable explanations even for larger models with short texts. AMV continues to show the best robustness across all models, with scores ranging from 0.0014 for TinyBERT to 0.0051 for DeBERTa-xlarge, indicating highly stable and consistent explanations regardless of model complexity.

% \textbf{Overall,} 
LIME and AMV are consistently the most robust XAI methods, providing stable and consistent explanations across both datasets and various models, regardless of text length and model complexity. Grad-CAM offers a balance of performance, providing good robustness across different models, particularly for larger models. LRP and SHAP are the least robust, with SHAP showing significant performance decreases for larger models and LRP exhibiting high variability and instability.

\textbf{Consistency}:
Table ~\ref{consitency} presents the quantitative results of the Consistency metric.
LIME demonstrates excellent consistency across both datasets, with scores improving as model complexity increases. For IMDB, scores range from 0.9665 for TinyBERT to 0.9895 for DeBERTa-xlarge, while for TSE, scores range from 0.7568 to 0.9556, indicating highly reliable explanations for both long and short texts. SHAP also exhibits strong consistency, with scores increasing from 0.9002 for TinyBERT to 0.9775 for DeBERTa-xlarge on IMDB, and from 0.8322 to 0.9324 on TSE, suggesting that SHAP's explanations become more reliable as model complexity increases.
LRP shows very good consistency, with scores improving from 0.9417 for TinyBERT to 0.9887 for DeBERTa-xlarge on IMDB, and from 0.8556 to 0.9713 on TSE, indicating stable and reliable explanations for more complex models. InputXGradient demonstrates strong consistency, with scores increasing from 0.9341 for TinyBERT to 0.9799 for DeBERTa-xlarge on IMDB, and from 0.7583 to 0.8596 on TSE, highlighting its effectiveness in providing consistent explanations for larger models.
Conversely, Grad-CAM shows poor consistency across both datasets, with negative scores ranging from -0.9593 for TinyBERT to -0.9749 for DeBERTa-xlarge on IMDB, and from -0.8259 to -0.9325 on TSE, indicating significant inconsistency in its explanations. In contrast, AMV achieves perfect consistency with scores consistently at 0.9999 across all models, regardless of complexity, providing identical explanations across all instances and models.

% \textbf{Overall}, 
AMV and LIME are the most robust XAI methods, consistently providing stable and reliable explanations across both datasets and various models, irrespective of text length and model complexity. SHAP, LRP, and InputXGradient offer a balance of performance with reasonable consistency, particularly for larger models. However, Grad-CAM remains the least consistent method, with significant variability and instability in its explanations.

\begin{table*}[ht]
\caption{The quantitative results of the combined weighted metrics scores on various XAI methods and encoder-based language models for IMDB and TSE datasets. Higher scores indicate better overall performance.}
\centering
\scriptsize
\begin{adjustbox}{max width=\textwidth}
\begin{tabular}{ccccccccccccccccccccccc}
\toprule
& \multicolumn{5}{c}{IMDB} & \multicolumn{5}{c}{TSE} \\
\cmidrule(lr){2-6} \cmidrule(lr){7-11}
 & TinyBERT  & BERTbase &  BERTlarge & XLM-R & DeBERTa\_xlarge & TinyBERT & BERTbase & BERTlarge & XLM-R & DeBERTa\_xlarge  \\
\midrule
LIME & \textbf{0.8862} & \textbf{0.8755} & \textbf{0.8797} & \textbf{0.8873} & \textbf{0.8611} & \textbf{0.7989} & \textbf{0.7468} & \textbf{0.7785} & \textbf{0.8572} & \textbf{0.8611} \\
SHAP & 0.7308 & 0.7507 & 0.7427 & 0.7504 & 0.7427 & 0.7308 & 0.7507 & 0.7427 & 0.7504 & 0.7516 \\
LRP & 0.7588 & 0.5686 & 0.7329 & 0.6741 & 0.6809 & 0.6779 & 0.6562 & 0.6677 & 0.7055 & 0.7285 \\
InputXGradient & 0.6626 & 0.6569 & 0.6741 & 0.6865 & 0.7062 & 0.6572 & 0.6457 & 0.6635 & 0.6859 & 0.6978 \\
Grad-CAM & 0.6621 & 0.2355 & 0.6906 & 0.6934 & 0.6965 & 0.1824 & 0.6817 & 0.6906 & 0.7197 & 0.7250 \\
AMV & 0.5796 & 0.5241 & 0.5637 & 0.5561 & 0.5442 & 0.5991 & 0.5920 & 0.5733 & 0.5593 & 0.5436 \\
\bottomrule
\end{tabular}
\end{adjustbox}
\label{CombinedMetrics}
\end{table*}

\textbf{Contrastivity}: Table ~\ref{contrastivity} presents the quantitative results of the Contrastivity metric. For both the IMDB and TSE datasets, LIME demonstrates strong contrastivity for smaller models, with scores of 0.7065 for TinyBERT on IMDB and 0.6863 on TSE. However, LIME's performance decreases as model complexity increases, with scores dropping to 0.5766 for DeBERTa-xlarge on IMDB and 0.5838 on TSE, suggesting that its ability to highlight contrasting features diminishes with larger models.
SHAP exhibits moderate contrastivity across both datasets, with scores such as 0.6449 for TinyBERT on IMDB and 0.6552 on TSE. However, as model complexity increases, SHAP's scores decrease, dropping to 0.5448 for DeBERTa-xlarge on IMDB and 0.5356 on TSE, indicating reduced effectiveness for larger models.
LRP shows varying performance on the IMDB dataset, with a lower score of 0.5797 for BERTbase but significantly higher scores for more complex models like DeBERTa-xlarge, which achieves 0.9371. LRP maintains strong contrastivity on the TSE dataset with high scores, such as 0.8540 for BERTlarge and 0.9367 for DeBERTa-xlarge, suggesting it effectively highlights contrasting features as model complexity increases.
InputXGradient displays good contrastivity for smaller models, with scores of 0.7488 for TinyBERT on IMDB and 0.7968 on TSE. However, its performance declines sharply with increasing model complexity, as seen in the scores of 0.3942 for DeBERTa-xlarge on IMDB and 0.3661 on TSE.
Grad-CAM demonstrates weak contrastivity across both datasets, with low scores such as 0.5689 for TinyBERT on IMDB and 0.6623 on TSE, further declining with more complex models like DeBERTa-xlarge, which scores 0.3641 on IMDB and 0.3371 on TSE, indicating less effective highlighting of contrasting features.
In contrast, AMV shows poor contrastivity across all models on both datasets, with particularly low scores of 0.1546 for TinyBERT and 0.0384 for DeBERTa-xlarge on IMDB, and similarly low scores on TSE, indicating minimal effectiveness in highlighting contrasting features.

LRP emerges as the most effective XAI method in terms of contrastivity, especially for complex models like DeBERTa-xlarge. This suggests that LRP is particularly adept at highlighting differences in model predictions based on contrasting features. LIME and SHAP offer a balance of performance, providing moderate contrastivity, although their effectiveness decreases as model complexity increases. Grad-CAM and AMV show poor contrastivity, with significant variability and lower scores across different models, indicating less reliable explanations for highlighting contrasting features.

\textbf{Overall}, The study highlights that no single XAI technique excels universally across all metrics and models. However, our rigorous evaluation process has identified some reliable XAI techniques. A model simplification-based approach, LIME, consistently performs well across multiple evaluation metrics, making it a reliable choice for generating explanations that align with human reasoning, and are robust and consistent. Despite its limitations in contrastivity, AMV, an attention mechanism approach, excels in robustness and consistency, making it suitable for applications where stability and reliability are paramount. A layer-wise relevance propagation approach, LRP, shows promise in contrastivity, particularly for complex models, indicating its potential for tasks requiring identifying contrasting features. Perturbation-based techniques (such as SHAP) and gradient-based techniques (such as InputXGradient) demonstrate moderate performance across all metrics and models.

\textbf{Combined Weighted-metrics Scores (CWS)}: A combined weighted metrics approach is employed to assess the performance of various XAI methods across different encoder-only language models. The evaluation is based on four key metrics: Human-reasoning Agreement (HA), Robustness (R), Consistency (Cn), and Contrastivity (Ct), each assesses various aspects of XAI performance. Higher scores (closer to 1) indicate better performance for Human-reasoning Agreement, Consistency, and Contrastivity. However, lower scores (closer to 0) are preferable for robustness as they reflect an excellent explanation of stability under perturbations. To align this with the other metrics, we normalize the Robustness score by subtracting it from 1, ensuring all metrics are positively oriented. Each metric is considered equally important, so they are all assigned an equal weight ($\omega$) of 0.25. However, if specific metrics are more critical to the evaluation context, larger weights can be assigned to those metrics accordingly. The CWS for each XAI method on each encoder-based language model is computed using the formula: 
\begin{equation}    
CWS = \omega_{HA} \cdot HA + \omega_{Cn} \cdot Cn + \omega_{Ct} \cdot Ct + \omega_{R} \cdot (1 - R)
\end{equation}

where $\omega_{HA}$ + $\omega_{Cn}$ + $\omega_{Ct}$ + $\omega_{R}$ =1, and $\omega \geq 0. $

CWS provides a comprehensive evaluation, with higher scores indicating the superior overall performance of XAI techniques across different encoder-based language models, as shown in Table ~\ref{CombinedMetrics}. Based on combined weighted metric scores, 
LIME, a model simplification-based technique, consistently demonstrates strong performance across all models on the IMDB and TSE datasets. SHAP, a perturbation-based method, shows balanced and reliable results across models. The LRP approach exhibits more variability in performance, with significant fluctuations depending on the model. InputXGradient, a gradient-based method, maintains mid-range scores, offering consistent reliability but falling short of top-tier performance. Grad-CAM, also a gradient-based method, shows significant variability, with performance varying greatly across models. Finally, AMV consistently scores modestly across both datasets, suggesting it may be less effective than other XAI methods.

LIME aligns more effectively with human reasoning than other XAI methods because it uses localized, interpretable approximations tailored to each prediction instance. By selectively isolating and analyzing the influence of specific words or tokens, LIME approximates complex model behavior in a way that resonates with human judgment. With their sophisticated attention mechanisms that process multi-dimensional relationships across tokens, transformer-based models are often opaque and challenging to interpret globally. LIME’s localized focus effectively addresses this issue by concentrating only on the most relevant tokens within each instance, avoiding the need to explain the entire model's behavior.
In contrast, other explainability methods, such as gradient-based techniques like Grad-CAM and InputXGradient, often lack the strong alignment with human reasoning that LIME provides. Gradient-based methods depend on gradients or feature attributions derived from the entire model behavior, which may introduce noise in highly parameterized transformer models. These methods frequently struggle to capture localized patterns and may highlight features that do not align with human interpretative strategies, especially as model complexity increases.

\section{Limitation and Future Work}
Due to experimental complexity, our study is limited to selected metrics, encoder-based language models, XAI techniques, and classification tasks. Future research could broaden the scope by incorporating a wider range of XAI techniques, exploring more diverse and complex transformer-based models such as  LLaMA, including under-resourced languages, and examining a broader set of downstream tasks. This extended work would allow for more refined evaluation metrics and increased applicability in real-world contexts. Furthermore, addressing the computational complexity of XAI techniques will be crucial to improving the scalability and feasibility of our evaluation framework for large-scale applications.

\section{Conclusion} \label{sectionConclusion}

Our proposed comprehensive evaluation framework, with a detailed set of metrics, serves as a structured approach to assess the effectiveness of various explainability techniques applied to encoder-based language models. This systematic evaluation, which rigorously tests these techniques across key evaluation metrics such as Human-reasoning Agreement, Robustness, Consistency, and Contrastivity, is a significant step forward. By evaluating diverse datasets and models, we provide an in-depth analysis of how well each technique aligns with human judgment, remains robust under perturbations, provides consistent explanations, and highlights contrasting features. Our findings indicate that model simplification approaches like LIME perform well across multiple metrics, regardless of model complexity. Although it has limitations in contrastivity, the attention mechanism approach (e.g., AMV) excels in robustness and consistency metrics, making it ideal for applications requiring stability and reliability. The layer-wise relevance propagation (LRP) technique shows strong potential in contrastivity, particularly for complex models, suggesting its potential advantage in tasks that require identifying contrasting features. Perturbation-based techniques (e.g., SHAP) and gradient-based techniques (e.g., InputXGradient) demonstrate moderate performance across all metrics and models. This evaluation enhances our understanding of the strengths and limitations of current XAI techniques in encoder-based language models. It lays a foundation for future research to improve the reliability of explanations provided by explainability techniques in language models in real-world applications.

\bibliographystyle{elsarticle-num}
\bibliography{References.bib}

% \clearpage
\appendix

\section{Human-reasoning Agreement Example}
\label{Example}
We proposed four evaluation metrics: Human-reasoning Agreement, Consistency, Robustness, and Contrastivity. In this framework, human annotators rank only the most important words, while precision and rel (k) are computed automatically without human intervention.

The primary objective of the Human-reasoning Agreement (HA) metric is to rigorously evaluate XAI explanations, ensuring they align accurately with human judgment, particularly in safety-critical domains such as healthcare, finance, autonomous vehicles, aerospace, nuclear energy, defense, transportation safety, and legal compliance. HA’s restricted ranking approach is crucial for achieving exact interpretability and alignment with human priorities in these fields, where misinterpreting AI predictions can lead to severe consequences. HA ensures that XAI explanations support reliable and safe decision-making by focusing on strict ranking alignment, reducing the risk of errors, and enhancing the AI model’s capacity to operate within strict safety and regulatory standards.

\subsection*{Average Precision (AP) for a Single Instance}
\begin{equation} \label{equ.AP}
AP = \frac{\sum_{k=1}^{n} (P(k) \times rel(k))}{\text{Number of relevant tokens (n)}} 
\end{equation}

where:
\begin{itemize}
    \item $k$ is the rank in the sequence of retrieved relevant words,
    \item $n$ is the total number of relevant words (as determined by human annotations),
    \item $P(k)$ is the precision at rank $k$,
    \item $\text{rel}(k)$ is an indicator function that equals 1 if the word at rank $k$ matches the human annotation; otherwise, it equals 0.
\end{itemize}

\subsection*{Mean Average Precision (MAP) across Multiple Instances}
\begin{equation}
\text{MAP} = \frac{1}{N} \sum_{n=1}^{N} \text{AP}_n
\end{equation}

where:
\begin{itemize}
    \item $N$ is the total number of instances evaluated,
    \item $\text{AP}_n$ is the Average Precision for the $n$-th instance.
\end{itemize}

\section*{Example }
(Note: This example is not actual program output; it is provided solely to illustrate how this metric functions.)\\
Input instance: " \textbf{The movie was absolutely fantastic, fascinating, and delightful.}"

\subsection*{Step 1: Human Annotation and Ranking (Gold Standard)}
Human annotators rank the words by importance for positive sentiment:\\
\{\text{fantastic}, \text{fascinating}, \text{absolutely}, \text{delightful}, \text{movie}\}

\subsection*{Step 2: XAI Explanation and Ranking} We
rank important words based on the XAI explanation scores  as follows:\\
\{\text{fantastic}, \text{fascinating}, \text{absolutely},  \text{movie}, \text{delightful}\}

\subsection*{Step 3: Computation of Relevance $\text{rel}(k)$ and Precision $P(k)$}
\textbf{Rank 1} (XAI: “fantastic”, Human: “fantastic”)\\
\textbf{Relevance} $rel(1)$: "fantastic" matches in both lists, so $rel(1) = 1$.\\
\textbf{Precision} $P(1)$: Computed as the number of relevant words up to rank 1 divided by 1: $P(1) = \frac{1}{1} = 1$

\textbf{Rank 2} (XAI: “fascinating”, Human: “fascinating”)\\
\textbf{Relevance} $rel(2)$: "Fascinating" also matches, so $rel(2) = 1$.\\
\textbf{Precision} $P(2)$: Computed as the number of relevant words up to rank 2 divided by 2: $P(2) = \frac{2}{2} = 1$

\textbf{Rank 3} (XAI: “absolutely”, Human: “absolutely”)\\
\textbf{Relevance} $rel(3)$: "Absolutely" matches, so $rel(3) = 1$.\\
\textbf{Precision} $P(3)$: Computed as the number of relevant words up to rank 3 divided by 3: $P(3) = \frac{3}{3} = 1$

\textbf{Rank 4} (XAI: “movie”, Human: “delightful”)\\
\textbf{Relevance} $rel(4)$: "Movie" does not match with "delightful," so $rel(4) = 0$.\\
\textbf{Precision} $P(4)$: Computed as the number of relevant words up to rank 4 divided by 4: $P(4) = \frac{3}{4} = 0.75$

\textbf{Rank 5} (XAI: “delightful”, Human: “movie”)\\
\textbf{Relevance} $rel(5)$: "Delightful" does not match with "movie," so $rel(5) = 0$.\\
\textbf{Precision} $P(5)$: Computed as the number of relevant words up to rank 5 divided by 5: $P(5) = \frac{3}{5} = 0.6$

\textbf{Summary}\\
\begin{center}
\begin{tabular}{|c|c|c|c|c|}
\hline
Rank & XAI Word & Human Word &  $rel(k)$ & $P(k)$ \\
\hline
1 & fantastic & fantastic & 1 & 1 \\
2 & fascinating & fascinating & 1 & 1 \\
3 & absolutely & absolutely & 1 & 1 \\
4 & movie & delightful & 0 & 0.75 \\
5 & delightful & movie & 0 & 0.6 \\
\hline
\end{tabular}
\end{center}

\textbf{Average Precision (AP):}\\
\begin{align*}
AP &= \frac{(1 \times 1) + (1 \times 1) + (1 \times 1) + (0.75 \times 0) + (0.6 \times 0)}{5} \\
   &= \frac{1 + 1 + 1 + 0 + 0}{5} = \frac{3}{5} = 0.6
\end{align*}

Thus, for this single instance, $AP = 0.6$.

\textbf{Mean Average Precision (MAP)}\\
If we evaluate multiple instances, MAP would be calculated as each instance's mean of AP scores. Since this example uses only one instance, $MAP = AP = 0.6$.

Hence, The higher the Average Precision (AP) and Mean Average Precision (MAP) scores, the stronger the alignment or agreement between the XAI explanations and human reasoning.

% \appendix
% begin{multicols}{1}
\onecolumn
\section{Sample XAI visualization Outputs}
\label{Visualization}
This appendix provides a comprehensive perspective on how different models and explainability methods handle sentiment analysis. we present sample visualizations of the decision-making process for sentiment predictions on different datasets, utilizing various explainability methods, including SHAP, LIME, InputXGradient, Grad-CAM, Attention Visualization, and Layer-wise Relevance Propagation (LRP). Each figure highlights the words contributing most significantly to the sentiment predictions, offering insight into how models like TinyBERT, BERT\_base, BERT-large, XLMR, and DeBERTa interpret input text data.
For this visualization, we selected two sample texts from the IMDB and TSE datasets, applying various models and explanation methods to compare how each interprets and attributes sentiment. This visualization comparison offers insight into model behavior and reveals how different explainability techniques illustrate the model decision-making process.
% \end{multicols}

\begin{figure}[ht]
\centering
\includegraphics[width=\textwidth]{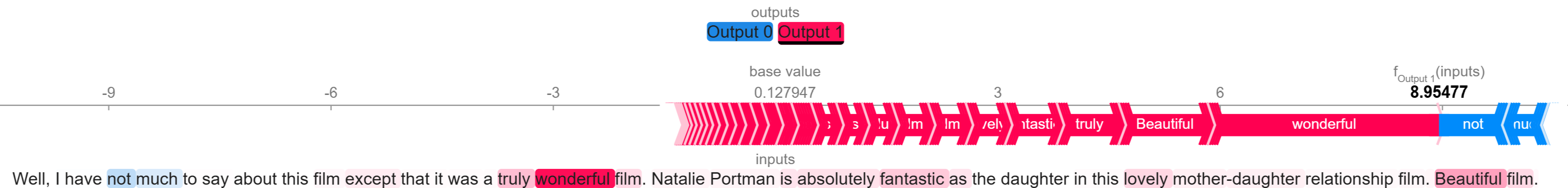}
% \vspace{10pt} % Adjust as needed
\captionsetup{justification=centering}
\begin{minipage}{\textwidth}
\vspace{5pt}
    \caption{\textbf{SHAP} explanation of BERT\_base model's IMDB movie sentiment prediction. Positive words are highlighted in red and negative words in blue.}
    % \label{fig:lime_bert}
\end{minipage}
\end{figure}

% ////

\begin{figure}[ht]
\centering
\includegraphics[width=\textwidth]{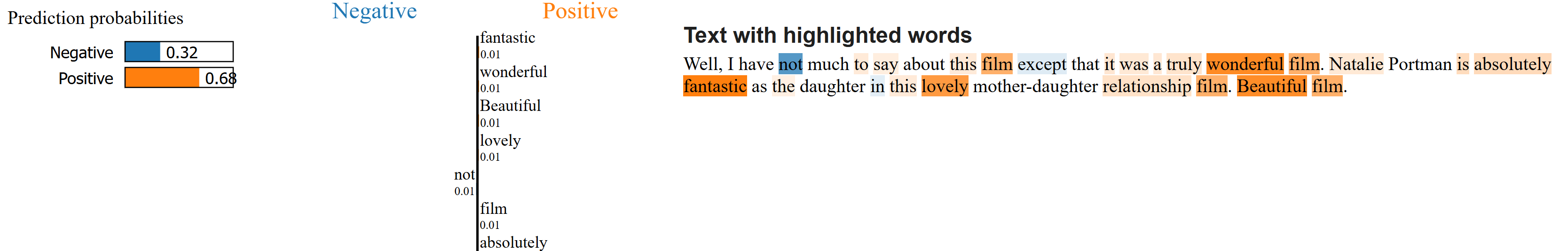}
% \vspace{5pt} % Adjust as needed
\captionsetup{justification=centering}
\begin{minipage}{\textwidth}
    \caption{\textbf{LIME} explanation of BERT\_large model's IMDB movie sentiment prediction. Positive words are highlighted in orange and negative words in blue.}
    % \label{fig:lime_bert_large}
\end{minipage}
\end{figure}

\begin{figure}[ht]
\centering
\includegraphics[width=\textwidth]{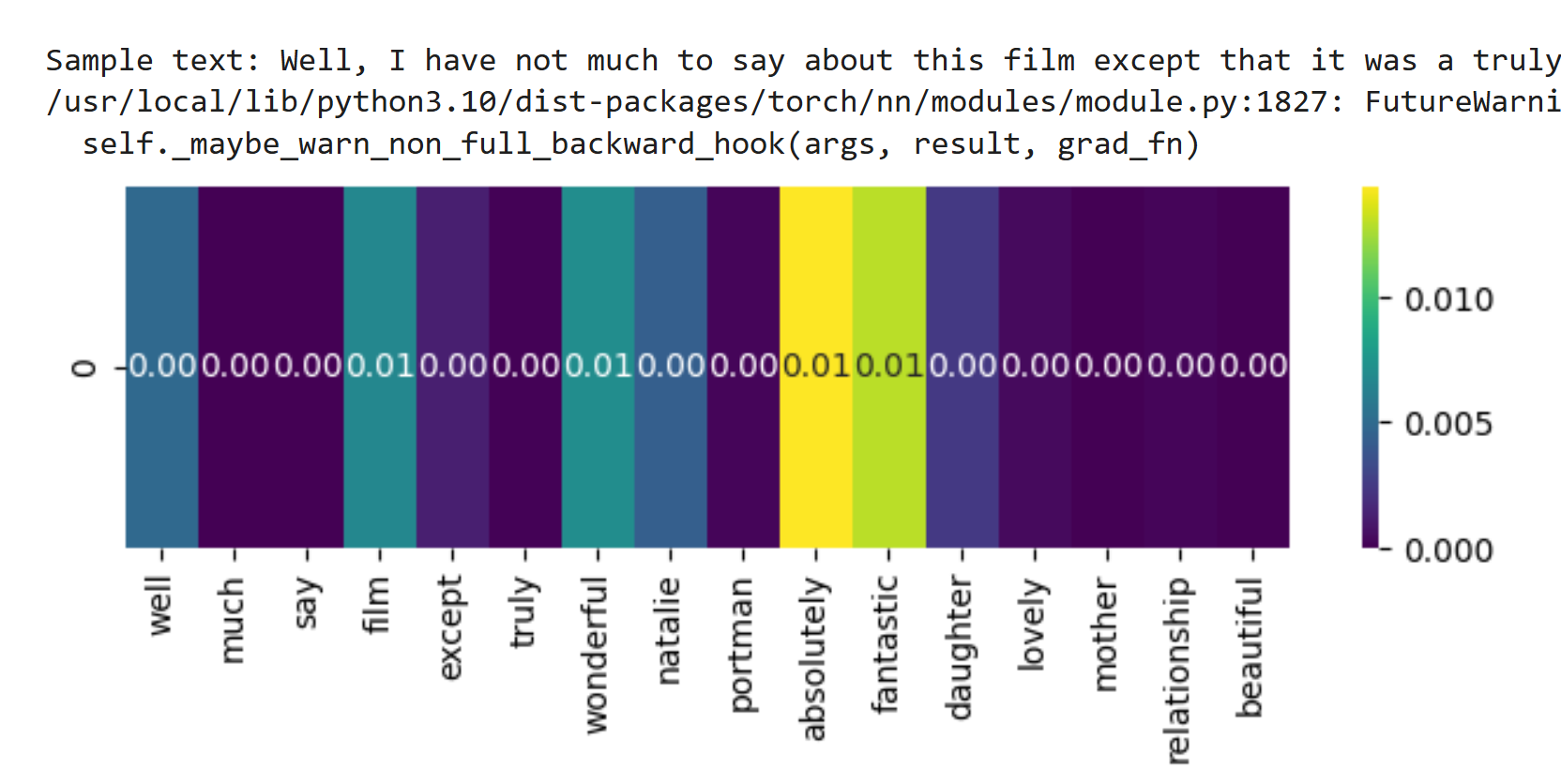}
% \vspace{5pt} % Adjust as needed
% \captionsetup{justification=centering}
\captionsetup{justification=raggedright, singlelinecheck=off}
\begin{minipage}{\textwidth}    
    \caption{ \textbf{Grad-CAM} explanation of DeBERTa model's positive sentiment prediction on IMDB movie review (Yellow indicates high importance and dark purple indicates less importance).}
    \label{fig:Grad_CAM}
\end{minipage}
\end{figure}

\begin{figure*}[ht]
\centering
\includegraphics[width=\textwidth]{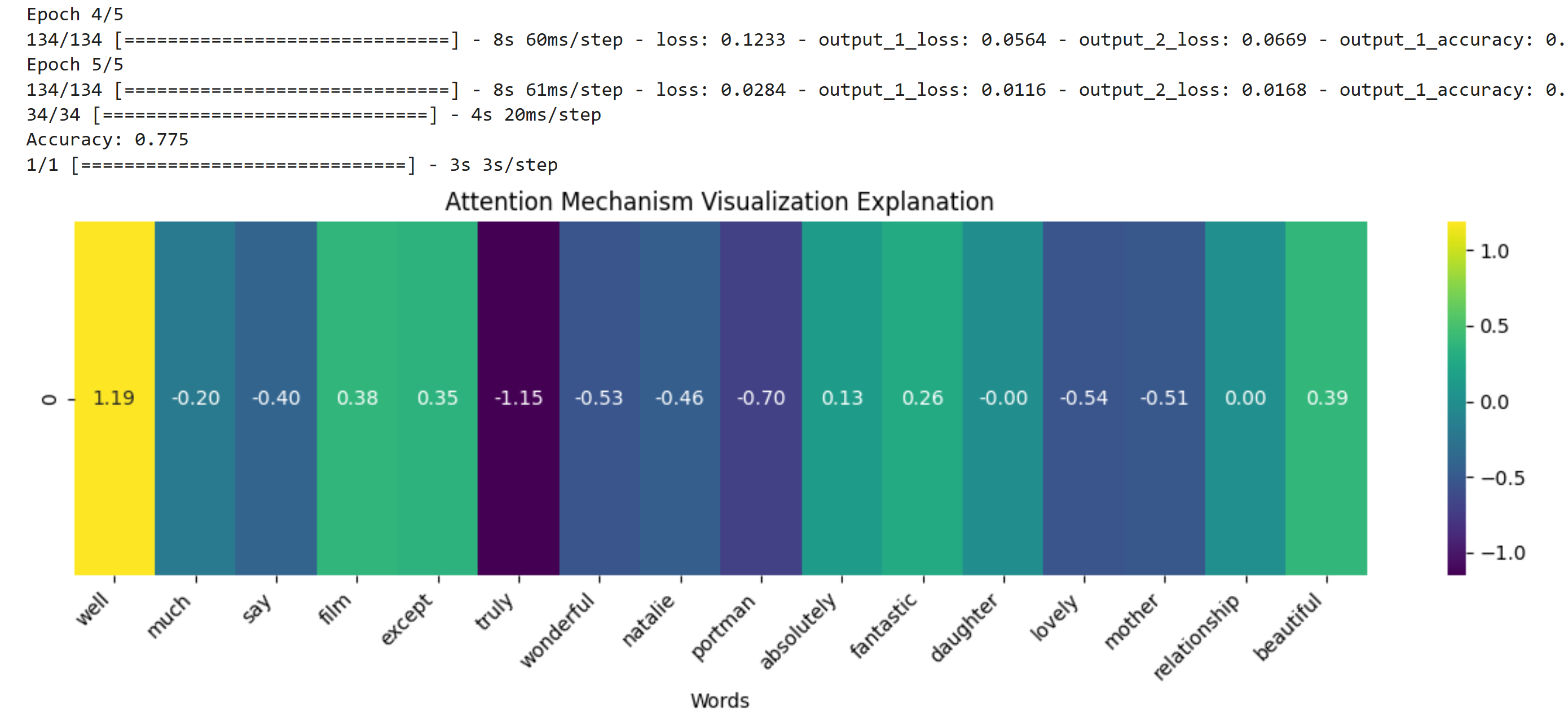}
% \vspace{5pt} % Adjust as needed
% \captionsetup{justification=centering}
\captionsetup{justification=raggedright, singlelinecheck=off}
\begin{minipage}{\textwidth}
    \caption{ \textbf{Attention Mechanism visualization} explanation of XLMR model's positive sentiment prediction on IMDB movie review (Yellow indicates high importance and dark purple indicates less importance).}
    \label{fig:AttXlmr}
\end{minipage}
\end{figure*}

% InputGrad
\begin{figure*}[ht]
\centering
\includegraphics[width=\textwidth]{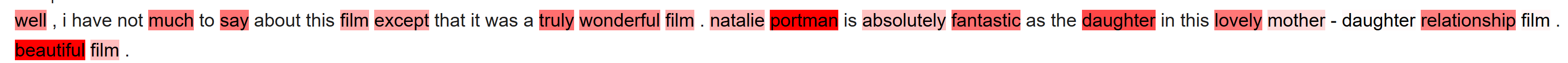}
% \vspace{5pt} % Adjust as needed
% \captionsetup{justification=centering}
\captionsetup{justification=raggedright, singlelinecheck=off}
\begin{minipage}{\textwidth}
\vspace{5pt}
    \caption{  \textbf{InputXGradient} explanation of TinyBERT model's positive sentiment prediction on IMDB movie review, highlighting the words that contribute most to the model's prediction. Stronger red intensity indicates higher relevance to the prediction, with words like 'beautiful,' 'fantastic,' 'wonderful,' and 'relationship' suggesting a positive sentiment.}
    \label{fig:inpugradient}
\end{minipage}
\end{figure*}
% //////////

% LIME
\begin{figure*}[ht]
\centering
\includegraphics[width=\textwidth]{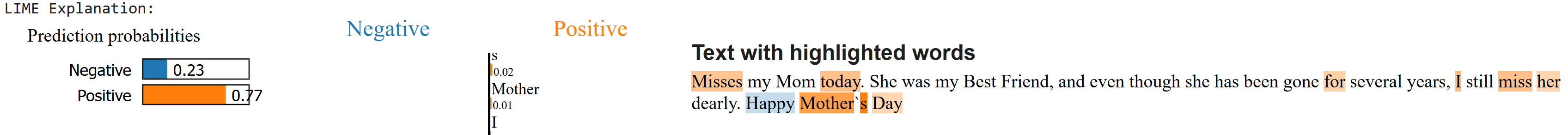}
% \vspace{5pt} % Adjust as needed
% \captionsetup{justification=centering}
\captionsetup{justification=raggedright, singlelinecheck=off}
\begin{minipage}{\textwidth}
    \caption{\textbf{LIME} explanation of BERT\_large model's TSE dataset sentiment prediction. Positive words are highlighted in orange, and negative words in blue.}
    % \label{fig:lime}
\end{minipage}
\end{figure*}

% ////////////////

% InputGrad
\begin{figure*}[ht]
\centering
\includegraphics[width=\textwidth]{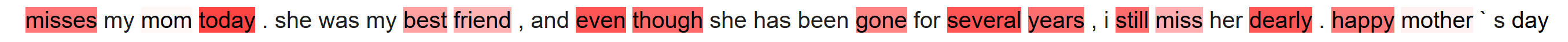}
\captionsetup{justification=raggedright, singlelinecheck=off}
\begin{minipage}{\textwidth}
\vspace{5pt}
    \caption{\textbf{InputXGradient} explanation of XLMR model's positive sentiment prediction on Tweet Sentiment Extraction (TSE) dataset, highlighting the words that contribute most to the model's prediction. Stronger red intensity indicates higher relevance to the prediction, with words like 'misses,' 'today,' 'even though,' 'gone,' 'several years,' 'still,' 'dearly,' and 'happy' suggesting a positive sentiment.}
    \label{fig:xlmr1}
\end{minipage}
\end{figure*}
% //////////

% LRP
\begin{figure*}[ht]
\centering
\includegraphics[width=\textwidth]{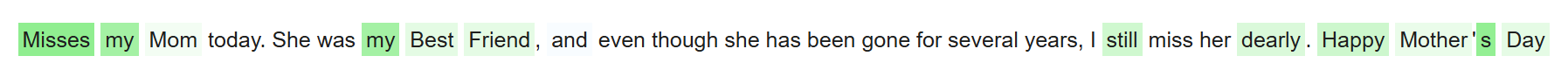}
\captionsetup{justification=raggedright, singlelinecheck=off}
\begin{minipage}{\textwidth}
    \vspace{5pt}
    \caption{ \textbf{LRP} explanation of BERT\_large model's TSE dataset sentiment prediction. Higher-importance words are highlighted with greater intensities of green, indicating their strong positive contribution to the BERT\_large model's prediction}
    \label{fig:bertLarge}
\end{minipage}
\end{figure*}

\end{document}